\pdfoutput=1

\documentclass[11pt]{article}

\usepackage[]{acl}

\usepackage{times}
\usepackage{latexsym}

\usepackage[T1]{fontenc}

\usepackage[utf8]{inputenc}
\usepackage{microtype}
\usepackage{amsmath}
\usepackage{amsfonts} 
\usepackage{graphicx}
\usepackage{booktabs}
\usepackage{multirow}
\usepackage{amssymb,paralist}
\usepackage{float}
\usepackage{pifont}

\newcommand{\xmark}{\ding{55}}%

\title{Open-Domain Conversational Question Answering with Historical Answers}

\author{Hung-Chieh Fang\footnotemark[1] \quad Kuo-Han Hung\thanks{\hspace*{0.5em}Equal contribution.}\quad Chao-Wei Huang\quad Yun-Nung Chen\\
  National Taiwan University, Taipei, Taiwan \\
  \texttt{\{b09902106,b09902120,f07922069\}@csie.ntu.edu.tw} \\ \texttt{y.v.chen@ieee.org} }

\begin{document}
\maketitle

\begin{abstract}
Open-domain conversational question answering can be viewed as two tasks: passage retrieval and conversational question answering, where the former relies on selecting candidate passages from a large corpus and the latter requires better understanding of a question with contexts to predict the answers. This paper proposes \textbf{ConvADR-QA} that leverages historical answers to boost retrieval performance and further achieves better answering performance. 
Our experiments on the benchmark dataset, OR-QuAC, demonstrate that our model outperforms existing baselines in both extractive and generative reader settings, well justifying the effectiveness of historical answers for open-domain conversational question answering.\footnote{The source code is available at \url{https://github.com/MiuLab/ConvADR-QA}.}
\end{abstract}

\section{Introduction}
Conversational information seeking and conversational question answering (CQA) are fundamental tasks of dialogue systems~\cite{gao-etal-2018-neural-approaches}.
The conversational agents are expected to serve as nature interfaces for users' information need, providing information and answers via multi-turn natural language interactions.
The multi-turn natural of CQA makes it challenging as the queries are contextualized, requiring the systems to resolve coreference and ambiguities.
With recent advances in language understanding and dialogue modeling, along with the curation of large-scale datasets, e.g., QuAC~\cite{choi-etal-2018-quac} and CoQA~\cite{reddy-etal-2019-coqa}, we have seen substantial progress in CQA.

While the state-of-the-art (SOTA) models have achieved performance comparable or even superior than human performance on QA and CQA datasets, this setting is highly limited as it requires the source document containing evidence to be given, which is unlikely the case in real-world scenarios.
To address this issue, researchers have expanded the scheme of CQA to an open-domain setting, where the document containing evidence must be retrieved from a large candidate pool~\cite{qu2020open}.
In the open-domain setting, there are usually millions of candidate documents, making the conventional method which jointly encodes the query and the document infeasible~\cite{chen-etal-2017-reading}.
The dominant technique to tackle the challenge is dense retrieval~\cite{karpukhin-etal-2020-dense,qu2020open,xiong2021approximate}, which encodes a query and  documents as dense representations separately and performs nearest neighbor search that is efficient and scalable to millions of documents.
It has been shown to outperform traditional sparse retrieval methods on multiple QA benchmarks.

However, applying dense retrieval for conversations may need to consider the dialogue context and structure, which is not trivial.
\citet{qu2020open} proposed ORConvQA to include previous questions in the same dialogue. where the context-dependent nature of questions is shown useful.
ConvDR~\cite{yu2021few} further improved the retrieval performance by knowledge distillation on reformulated questions with an ad-hoc teacher model.
Nevertheless, simply concatenating historical questions is suboptimal.
Our hypothesis is that rather than relying on the model to infer helpful knowledge from historical questions, we provide direct signals by adding historical answers to the input.
Hence, we propose \textbf{ConvADR-QA} (\textbf{Conv}ersational \textbf{A}nswer-aware \textbf{D}ense \textbf{R}etrieval) to leverage historical answers for better retrieval and then answering performance for open-domain CQA.


\section{Related Work}
\paragraph{CQA}
A unique challenge to CQA is that the questions are context-dependent.
Hence, most prior work focused on various history modeling techniques~\cite{huang2018flowqa,yeh-chen-2019-flowdelta,10.1145/3331184.3331341,ijcai2020-171}.
\citet{choi-etal-2018-quac} proposed to mark the previous answers in the passage by adding an answer embedding to the input embeddings.
\citet{bert_hae} extended this method to the large pre-trained language models.
However, \citet{chiang2020empirical} showed that prior conversational models do not fully understand the content, implying that CQA still needs further investigation.
While our method also leverages historical answers as additional input signal, our major contribution is that we apply this technique to dense retrieval instead of question answering for better practicality in a open-domain setting.

\vspace{-1mm}
\paragraph{Open-Domain QA}
Without a given target passage, most work for this task was built upon the dense retrieval framework for retrieving relevant passages for QA.
DPR~\cite{karpukhin-etal-2020-dense} first showed that dense retrieval outperforms sparse retrieval methods.
GAR~\cite{mao-etal-2021-generation} introduced pseudo relevance feedback by augmenting queries with generated texts.
RIDER~\cite{mao-etal-2021-reader} proposed a simple passage reranking method which promotes the passages containing the predicted answers.
While these methods consider the predicted answers, they aim at improving single-turn question answering.
We instead focus on enhancing model's ability on handling multi-turn conversational questions.

\vspace{-1mm}
\paragraph{Open-Domain CQA}
Researchers have put increasing attention on open-domain CQA with the TREC Conversational Assistance Track~\cite{dalton2020trec,dalton2021cast}.
However, these datasets have limited supervision, making dense retrieval hardly applicable due to its data-hungry nature.
\citet{qu2020open} introduced the first large-scale open-domain CQA data, OR-QuAC, by extending QuAC~\cite{choi-etal-2018-quac} to a open-domain setting.
They also proposed ORConvQA, a pipeline system with a DPR retriever and an extractive reader, as a baseline system.
ConvDR~\cite{yu2021few} proposed to reformulate questions into their context-independent rewrites with the CANARD dataset~\cite{elgohary-etal-2019-unpack}, then applied knowledge distillation using a ad-hoc teacher model.
Our method is built upon these two methods by incorporating historical answers to aid the retriever.
\citet{li2021graph} proposed a graph-guided retrieval method which constructs a graph using passages with historical answers and potential answers.
Our work does not introduce extra parameters and complex modeling, and we demonstrate that we can achieve better results with a simpler design for better practicality. The AllHistory strategy from TopiOCQA \cite{adlakha-etal-2022-topiocqa} is very similar to ours. However, their experimental setting is not realistic as they used ground truth answers as historical answers, which is corresponding to our oracle setting.

\section{ConvADR-QA}

\begin{figure}[t!]
\centering
\includegraphics[width=.95\linewidth]{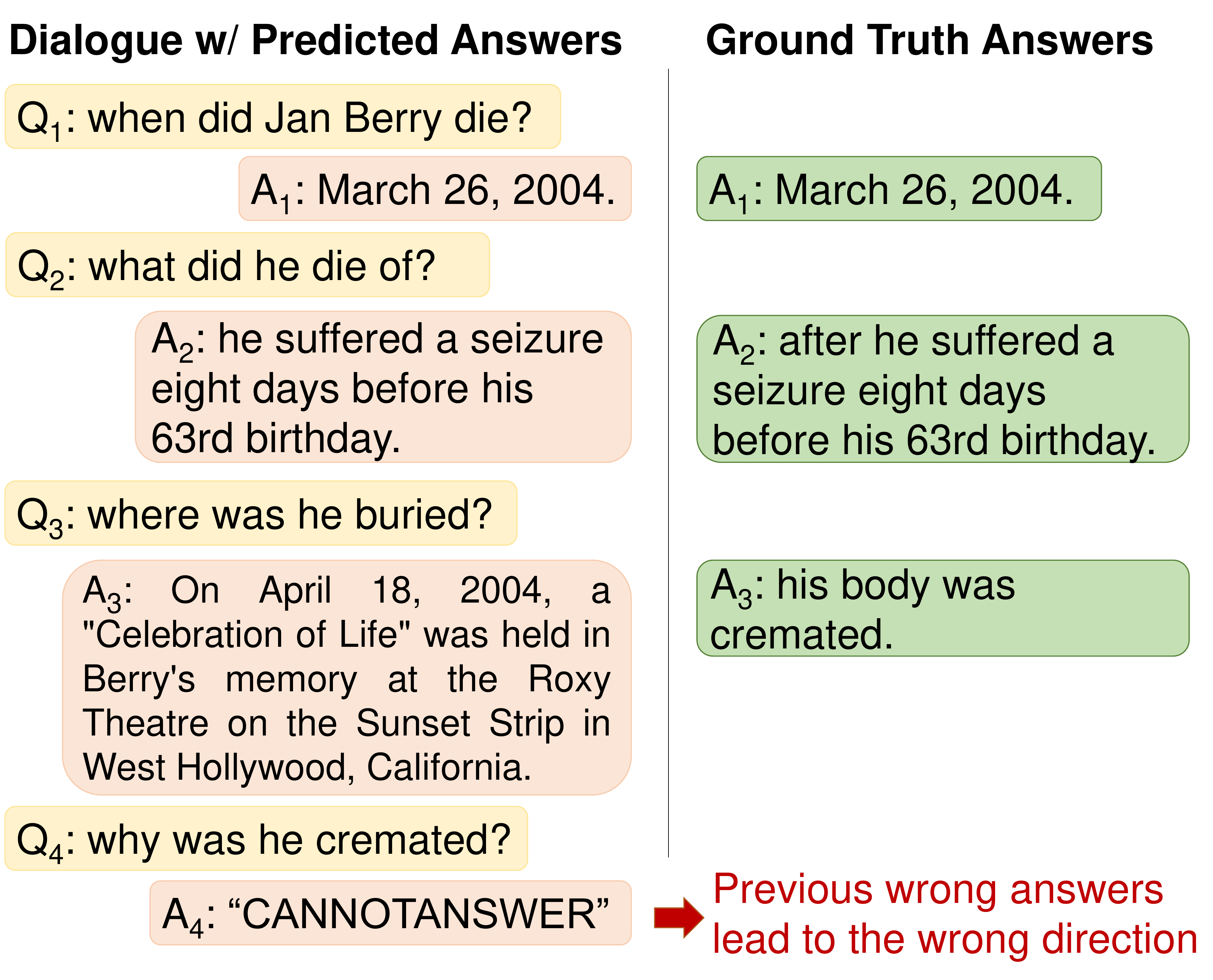} 
\vspace{-1.5mm}
\caption{Demonstration of how previous answers affect the quality of an answer.} 
\label{fig:demo}
\vspace{-3.5mm}
\end{figure}

\begin{figure*}[t]
\centering
\includegraphics[width=.95\linewidth]{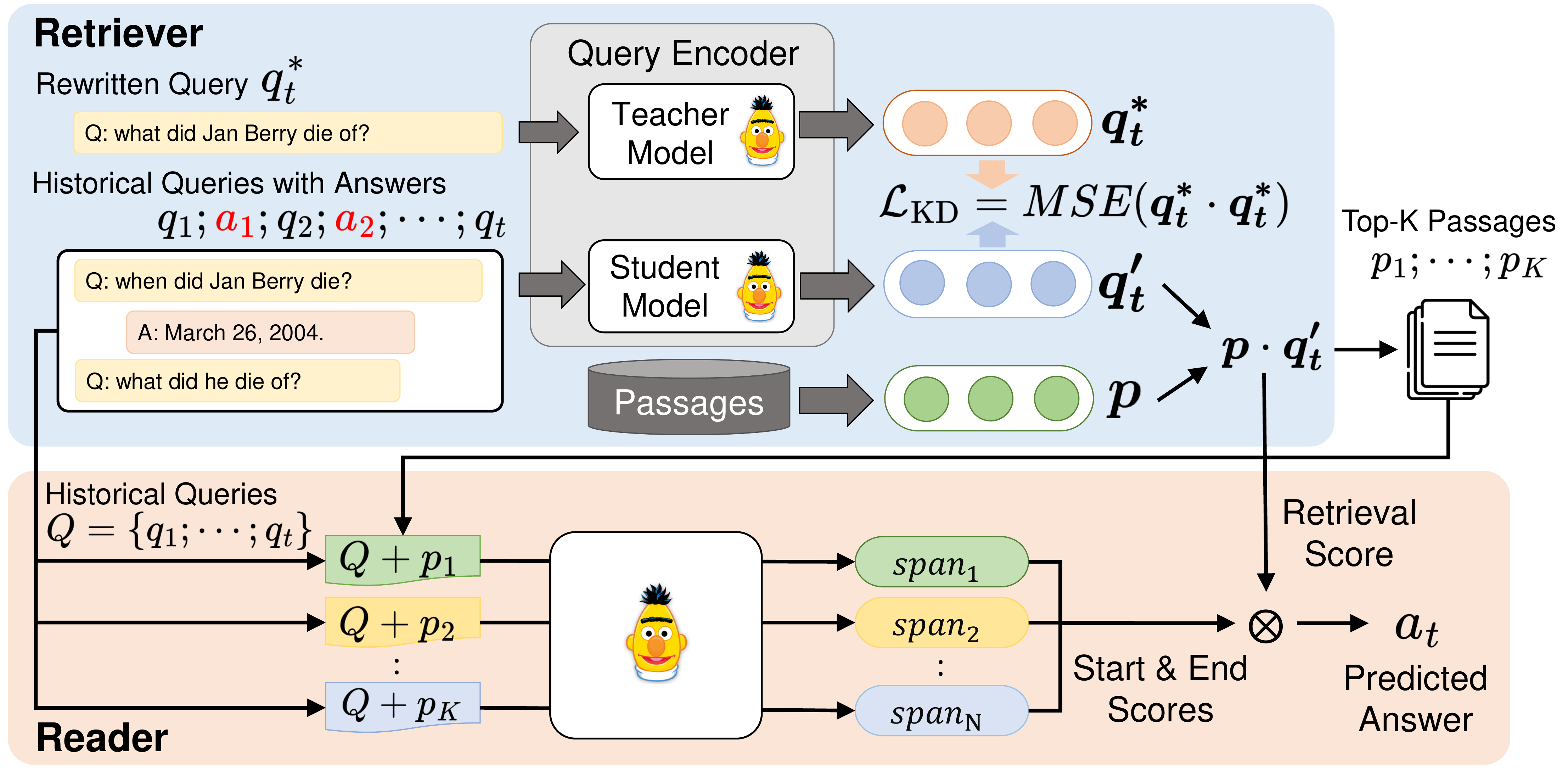} 
\vspace{-1.5mm}
\caption{Illustration of our proposed ConvADR-QA model.}
\label{fig:model}
\vspace{-3mm}
\end{figure*}

Let $C$ denote the passage collections with $N$ passages $\{p_i\}_{i=1}^{N}$, where $p_i$ can be viewed as a sequence of tokens $p_{i}^1,\dots ,p_{i}^l$. Given the $t$-th question $q_t$ and all historical questions $\{q_i\}_{i=1}^{t-1}$ in a conversation, the task of open-domain CQA is to predict $a_t$ from $C$. In an extractive setting, $a_t$ is a span $p_i^s,\dots ,p_i^e$ from a passage $p_i$.

The difficulty of open-domain CQA is that the current question usually requires context information from previous turns, which makes it harder for the system to capture the latent information compared with the open-domain QA task.
Previous work on open-domain conversational search addressed the problem by concatenating the current and historical questions \emph{without} answers~\cite{qu2020open, yu2021few}. Our motivation is that historical answers can also provide the important signal for the current question to obtain the answers
illustrated in Figure \ref{fig:demo}. 

To better leverage the historical answers for open-domain CQA, we propose ConvADR-QA illustrated in Figure \ref{fig:model}, which includes a retriever for obtaining relevant passages from a large collection and a reader for CQA.

\subsection{Retriever}
Following the prior work~\cite{karpukhin-etal-2020-dense, xiong2021approximate}, we apply a dense retrieval method, which has shown dominant performance over sparse ones.
Specifically, the model uses a dual-encoder architecture to map passages and questions to the same embedding space.
The input of our question encoder is the concatenation of historical questions and answers:
\[
    \boldsymbol{p} = E_{P}(p),
    \boldsymbol{q_k'} = E_{Q}(\{q_i, a_i\}_{i=1}^{k-1}; q_k).
\]
The retrieval score is then defined as the dot product of the passage embedding and the question embedding:
\[
    S_\text{rt}(q_k, p) = \boldsymbol{p} \cdot \boldsymbol{q_k'}.
\]
In the training process, each question contains one gold passage $p^+$ and a set of negative passages $P^-$, ConvADR-QA is then optimized using the negative log likelihood loss:
\[
        \mathcal{L}_{\text{NLL}} = -\log{\frac{e^{S_\text{rt}(q_k, p^+)}}{e^{S_\text{rt}(q_k, p^+)} + \sum_{p^- \in P^-}e^{{S_\text{rt}(q_k, p^-)}}}}.
\]
\subsection{Knowledge Distillation}
In conversational search, dense retrieval is challenging since the current question requires information from previous turns, which aggravate the discrepancy between question embeddings and passage embeddings. \citet{yu2021few} recently addressed the problem using a teacher-student framework to distill knowledge from an ad-hoc teacher model.
The input of the teacher model is a manually-rewritten context-independent query $q_k^{\star}$, and the knowledge distillation (KD) loss is defined as the mean square error (MSE) loss between the teacher's and the student's question embeddings:
\[
    \boldsymbol{p} = E'_{P}(p), 
   \boldsymbol{q_k^{\star}} = E'_{Q}(q_k^{\star}),
\]
\[
    \mathcal{L}_{\text{KD}} = \text{MSE}(\boldsymbol{q_k^{\star}}, \boldsymbol{q_k'}).
\]
 The retrieval loss of our multi-task learning setting is the sum of NLL loss and KD loss:
 \[
    \mathcal{L}_{\text{NLL}} + \mathcal{L}_{\text{KD}}.
 \]
\subsection{Reader}
The task of the reader is to extract a span from passages as the final answer. We use a standard BERT model for the machine comprehension task~\cite{kenton2019bert}. Given the $t$-th question $q_t$ and top-$K$ candidate passages $\{p_i\}_{i=1}^K$ retrieved by our retriever, the reader first extracts a span for each passage by choosing the highest score of start and end tokens. The score of the $m$-th token is defined as follows:
\[
    S_\text{start}^{[m]}(q_t; p) = W_\text{start}\text{BERT}(\{q_i\}_{i=1}^{t}; p)[m],
\]
\[
    S_\text{end}^{[m]}(q_t; p) = W_\text{end}\text{BERT}(\{q_i\}_{i=1}^{t}; p)[m],
\]
\[
    S_\text{rd}(q_t; p) = \max_{m1, m2}[S_\text{start}^{[m1]}(q_t; p)+S_\text{end}^{[m2]}(q_t; p)].
\]

We choose the final answer by multiplying the retriever score $S_\text{rt}$ and the sum of start/end token score as the reader score $S_\text{rd}$:
\[
S(q_t, p) = S_\text{rt}(q_t, p) \cdot S_\text{rd}(q_t; p).
\]

\begin{table*}[t!]
\centering 
\resizebox{0.99\linewidth}{!}{
\begin{tabular}{clc|ccc|ccc}
\toprule
& \multirow{2}{*}{\bf Method} & \bf Historical & \multicolumn{3}{c|}{\bf Retrieval} & \multicolumn{3}{c}{\bf Answering}\\ 
& & \bf Answers & \bf MRR@5 & \bf R@5 & \bf MAP@10 & \bf HEQ-Q & \bf HEQ-D & \bf F1\\
\midrule \midrule
\multirow{5}{*}{\rotatebox[origin=c]{90}{Extractive}} & ORConvQA & \xmark & 31.3 & 31.4 & - & 24.10 & 0.60 & 29.4 \\
& Graph-Guided & \small predicted & 35.1 & 36.7 & - & 30.30 & 1.00 & 33.4 \\
& ConvDR$\rightarrow$Reader & \xmark & 61.6 & 75.0 & 60.7 & 29.92 & 0.78 & 36.2 \\
\cmidrule{2-9}
& ConvADR-QA (Reader) & \small predicted & \textbf{66.8} & \textbf{77.9} & \textbf{64.6} & \textbf{32.11} & \textbf{1.16} & \textbf{38.4} \\
& ConvADR-QA (Reader) & \small gold & \it 74.5 & \it 82.5 &\it 71.7 & \it 35.69 & \it 1.03 & \it 42.3 \\ \midrule
\multirow{4}{*}{\rotatebox[origin=c]{90}{Generative}} & RAG & \xmark & 29.9 & 30.8 & 28.5 & 21.98 & 0.25 & 26.1 \\
& ConvDR$\rightarrow$FiD & \xmark & \bf 61.6 & 75.0 & 60.7 & 27.21 & 0.86 & 31.5\\
\cmidrule{2-9}
& ConvADR-QA (FiD) & \small predicted & 60.9 & \bf 76.2 & \bf 62.9 & \bf 28.76 & \bf 1.04 & \bf 33.6 \\
& ConvADR-QA (FiD) & \small gold & \it 74.5 & \it 82.5 & \it 71.7 & \it 30.83 & \it 0.91 & \it 35.1 \\
\bottomrule
\end{tabular}}
\vspace{-1mm}
\caption{Performance on OR-QuAC (\%). Best results are marked in {\bf bold}. Oracle results are in {\it italic}.}
\label{tab:convdr} 
\vspace{-3mm}
\end{table*}

\section{Experiments}
We conduct the experiments on an open-domain CQA benchmark: OR-QuAC~\cite{qu2020open}.
OR-QuAC is an open-domain conversational retrieval dataset that aggregates three existing datasets: (1) the QuAC dataset~\cite{choi-etal-2018-quac} which contains 14K information-seeking QA dialogs, (2) the CANARD dataset~\cite{elgohary-etal-2019-unpack} which rewrites context-dependent queries to self-contained questions based on QuAC, and (3)  the Wikipedia corpus dump from 10/20/2019 which extends QuAC to the open-domain setting. The experimental setting is detailed in Appendix~\ref{app:exp}.

Following \citet{yu2021few}, we use three commonly used metrics, MRR@5, Recall@5 and MAP@10, to evaluate the retrieval performance. In addition, we use word-level F1 and human equivalence score (HEQ) provided by the QuAC challenge
to evaluate the overall performance of our system. The definitions of above metrics are detailed in Appendix~\ref{app:eval}.

\subsection{Baselines}
We compare our model with recently proposed baselines for open-domain CQA, ORConvQA~\cite{qu2020open}, Graph-Guided~\cite{li2021graph}, and RAG~\cite{NEURIPS2020_6b493230}  for both extractive and generative settings.
\begin{itemize}
    \item \textbf{ORConvQA}: It is an end-to-end system for the open-domain CQA task, which includes a retriever, a reranker and a reader. The retriever use the dense retrieval method where the input of the query encoder is the concatenation of the current and historical questions. 
    \item \textbf{Graph-Guided}: \citet{li2021graph} proposed a graph-guided retrieval method that models the relations among answers across conversational turns, which is the first work attempting at utilizing historical answers for open-domain CQA.
This model utilizes a graph built from the hyperlink-connected passages containing historical answers to better retrieve relevant passages.
 \item \textbf{RAG}: It is a generation model that can access to pre-trained parametric memory and non-parametric memory like wikipedia. It has shown good performance on the open-domain QA task, we further adapt it to the open-domain CQA task by doing the following modifications: (1) finetuning the base model, where the input of the question encoder is the concatenation of the current and historical questions, (2) using passages from OR-QuAC as our knowledge source (non-parametric memory). 
\end{itemize}


In addition to the existing open-domain CQA approaches, we further implement two baseline where we use ConvDR as the retriever model.
ConvDR is a conversational dense retriever, which uses the few-shot strategy to mimic the embeddings of manual oracle queries from an ad hoc dense retriever. It is also the current SOTA model in the retrieval stage.
We adopt it to open-domain CQA by enabling it with QA capability using two existing models to generate answers: (1) Reader of ORConvQA~\cite{qu2020open}, which adapts a BERT-based extractive QA model to a multi-document setting, (2) FiD~\cite{izacard-grave-2021-leveraging}, which uses a sequence-to-sequence model to generate the answer given the input is the question and retrieved passages, which has shown great performance at combining evidences from multiple passages. 

\subsection{Results}
Table \ref{tab:convdr} summarizes our experimental results. 
It is obvious that our proposed ConvADR-QA outperforms almost all existing baseline models in both retrieval and answering stages, achieving new SOTA performance of open-domain CQA.
We can observe that in both extractive and generative QA settings, our model which leverages predicted answers achieves better performance over the one without answers. 
Moreover, the graph-guided approach also utilizes historical answers in a more complex way, but performs worse than our ConvADR-QA, demonstrating that our model leverages answer signal more effectively.
We also report the oracle results using gold historical answers. It shows that the model with gold answers outperforms the one with predicted answers in most of the metrics except HEQ-D.
Note that the oracle results can be viewed as the upper bound of our method, as the gold answers are not available during inference. 
The results well justify our hypothesis that historical answers are informative for open-domain CQA.

Notably, we can notice that the quality of predicted answers can significantly affect the retrieval performance.
Our experiment shows that MRR@5 drops when using FiD as the reader, demonstrating that a QA model with weak performance could potentially hurt retrieval performance.
Our hypothesis is that due to its lower answering quality, the errors would propagate through the conversation and mislead the retriever, indicating that further improvement on reader performance could also improve the retrieval performance of our method.
In sum, the experimental results show the effectiveness of our model for open-domain CQA in both extractive and generative settings. 
An example is presented in Table~\ref{tab:qualitative} for qualitative analysis, where it can be shown that the previous answers affect the following prediction results.
More analysis can be found in Appendix \ref{app:qualitative}.


\subsection{Error Propagation Analysis}
\begin{figure}[t!]
\centering
\includegraphics[width=.98\linewidth]{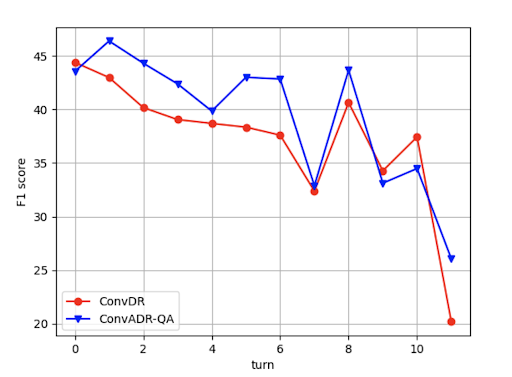} 
\vspace{-2mm}
\caption{Analysis on accuracy against number of turns.} 
\label{fig:error_prop}
\vspace{-1mm}
\end{figure}

To inspect the impact of the errors propagated through the conversation and reduce the robustness. we conduct analysis on accuracy against number of turns in Figure~\ref{fig:error_prop}.
It shows that the benefits of adding the answers outweigh the error propagation, where ConvADR-QA outperforms ConvDR in earlier turns, which tends to drop as the dialogue gets too long.
It implies that the issue about error propagation still have a large room for improvement.

\begin{table}[t!]
\centering
\small
\begin{tabular}{p{7.4cm}}
\toprule
\textbf{Q1}: What is Roberto Mangabeira Unger,'s programmatic thought? \\
\textbf{A1}: Key in Unger's thinking is the need to re-imagine social institutions before attempting to revise them. \\
\textbf{ConvDR}: The beginning of Unger's academic career began with the books Knowledge and Politics and Law in Modern Society,\\ 
\textbf{ConvADR-QA}: Key in Unger's thinking is the need to re-imagine social institutions before attempting to revise them.\\ 
 \midrule
 \textbf{Q2}: Can you explain the mechanism of thinking? \\
   \textbf{A2}: In building this program, however, we must not entertain complete revolutionary overhaul, lest we be plagued by three false assumptions: \\
 \textbf{ConvDR}: CANNOTANSWER\\ 
  \textbf{ConvADR-QA}: In building this program, however, we must not entertain complete revolutionary overhaul, lest we be plagued by three false assumptions:\\ 
 \midrule
  \textbf{Q3}: What are the three false assumptions? \\
  \textbf{A3}: Typological Fallacy: \\ 
  \textbf{ConvDR}: Unger finds three weaknesses that crippled the theory: foremost, the theory claimed that equilibrium would be spontaneously generated in a market economy.\\ 
  \textbf{ConvADR-QA}: Typological Fallacy: the fallacy that there is closed list of institutional alternatives in history, such as ``feudalism'' or ``capitalism''.\\ 
\bottomrule
\end{tabular}
\vspace{-1mm}
\caption{Qualitative analysis.}
\label{tab:qualitative} 
\vspace{-2mm}
\end{table}

\section{Conclusions}
This work introduces \textbf{ConvADR-QA}, an open-domain CQA model that leverages historical answers.
The experiments on a benchmark dataset demonstrate that our proposed method outperforms all baselines for both retrieval and answering performance.
Our results justify not only the importance of historical answers in a conversation but also the generalizability to different types of readers.

\section*{Acknowledgements}
We thank reviewers for their insightful comments. 
This work was financially supported from Google and the Young Scholar Fellowship Program by Ministry
of Science and Technology (MOST) in Taiwan,
under Grants 111-2628-E-002-016 and 111-2634-F-002-014.

\bibliography{anthology,custom}
\bibliographystyle{acl_natbib}

\appendix

\begin{table*}[t!]
\centering
\small
\begin{tabular}{lp{13cm}}
\toprule
\textbf{Example 1} \\
\midrule
\textbf{Q1} & What was unique about Joe Greene's attitude? \\
\textbf{A1} & ferocious and intimidating. \\
\textbf{ConvDR} & Greene's nickname remained popular due to his exploits on the playing field, where he was described as \underline{ferocious and intimidating}.\\ 
\textbf{ConvADR-QA} & Greene's nickname remained popular due to his exploits on the playing field, where he was described as \underline{ferocious and intimidating}.\\ 
\midrule
\textbf{Q2} & Did he intimidate other players?\\
\textbf{A2} & He instilled fear in opponents with the intensity of his play. \\ 
\textbf{ConvDR} & Lynn Swann, a wide receiver, considered Greene a mentor. \\
\textbf{ConvADR-QA} & \underline{He instilled fear in opponents with the intensity of his play.}\\
\midrule
\textbf{Q3} & Who did he play for?\\
\textbf{A3} & Pittsburgh \\
\textbf{ConvDR} & Houston Oilers\\
\textbf{ConvADR-QA} & Houston Oilers\\
\midrule
\textbf{Q4} & Did his teams do well? \\
\textbf{A4} & CANNOTANSWER \\
\textbf{ConvDR} & Greene was named the captain of the defense in 1977. \\
\textbf{ConvADR-QA} & \underline{CANNOTANSWER}\\
\midrule
\textbf{Q5} & What else should I know about his playing style? \\
\textbf{A5} & In his early years with the Steelers, Greene was at times uncontrollable, and often let his temper get the best of him. \\
\textbf{ConvDR} & Greene was named the captain of the defense in 1977.\\
\textbf{ConvADR-QA} & \underline{In his early years with the Steelers, Greene was at times uncontrollable, and often let his temper get the} \underline{best of him.}\\
\midrule
\multicolumn{2} {l} {\textbf{Example 2}} \\
\midrule
\textbf{Q1} & What is Roberto Mangabeira Unger,'s programmatic thought? \\
\textbf{A1} & Key in Unger's thinking is the need to re-imagine social institutions before attempting to revise them. \\
\textbf{ConvDR} & The beginning of Unger's academic career began with the books Knowledge and Politics and Law in Modern Society,\\ 
\textbf{ConvADR-QA} & \underline{Key in Unger's thinking is the need to re-imagine social institutions before attempting to revise them.}\\ 
 \midrule
 \textbf{Q2} & Can you explain the mechanism of thinking? \\
 \textbf{A2} & In building this program, however, we must not entertain complete revolutionary overhaul, lest we be plagued by three false assumptions: \\
 \textbf{ConvDR} & CANNOTANSWER\\ 
  \textbf{ConvADR-QA} & \underline{In building this program, however, we must not entertain complete revolutionary overhaul, lest we be} \underline{plagued by three false assumptions:}\\ 
 \midrule
  \textbf{Q3} & What are the three false assumptions? \\
  \textbf{A3} & Typological Fallacy: \\
  \textbf{ConvDR} & Unger finds three weaknesses that crippled the theory: foremost, the theory claimed that equilibrium would be spontaneously generated in a market economy.\\ 
  \textbf{ConvADR-QA} & \underline{Typological Fallacy:} the fallacy that there is closed list of institutional alternatives in history, such as ``feudalism'' or ``capitalism''.\\ 
\bottomrule
\end{tabular}
\vspace{-2mm}
\caption{The comparison between ConvDR and ConvADR-QA.}
\label{tab:qualitative} 
\end{table*}

\section{Reproducibility}
\label{app:exp}
Our source code and the trained model was published at GitHub together with running scripts for better reproducibility. 
All models are trained with 2 Nvidia Quadro P6000. For the retriever, we set the training batch size to 4, the number of epochs to 3, and the learning rate to 1e-5. For the reader, we set the training batch to 2, the number of epochs to 3, the max sequence length to 512, the max question length to 125 and the learning rate to 3e-5.

\section{Evaluation Metrics}
\label{app:eval}

We use following metrics for evaluating our proposed model.
\begin{itemize}
\item \textbf{MRR}: Reciprocal Rank (RR) calculates the reciprocal of the rank where the first relevant passage was retrieved, and MRR averages the reciprocal rank across all questions.
\item \textbf{Recall}: The proportion of the questions that the answer is in the retrieved passages.
\item \textbf{MAP}: Mean Average Precision is the mean of the average precision scores for each question.
\item \textbf{F1}: It measures the overlap of the predicted answer span
and the ground truth answer span at the word level.
\item \textbf{HEQ}: HEQ-Q measures the percentage of questions where the system matches or surpasses human performance in terms of F1 score.
HEQ-D measures the percentage of dialogues in which all questions have an HEQ-Q = 1.
\end{itemize}

%
\section{Qualitative Study}
\label{app:qualitative}
The results generated by ConvDR and ConvADR-QA are presented in Table \ref{tab:qualitative}, where the underline texts indicate the spans appearing in the ground truth answers. It can be found that historical answers can be an important signal to obtain the answers.

\end{document}